\documentclass{article}

\usepackage[main, final]{neurips_2026}

\usepackage[utf8]{inputenc} 
\usepackage[T1]{fontenc}    
\usepackage{hyperref}       
\usepackage{url}            
\usepackage{booktabs}       
\usepackage{amsfonts}       
\usepackage{nicefrac}       
\usepackage{microtype}      
\usepackage{xcolor}         

\usepackage{graphicx} 
\usepackage{amsmath}
\usepackage{amssymb}
\usepackage{mathtools}
\usepackage{amsthm}
\usepackage{subcaption}
\usepackage{multirow}

\usepackage{colortbl} 
\usepackage{xcolor}
\usepackage{tabularx} 
\usepackage{multirow}
\usepackage[utf8]{inputenc}

\usepackage{subcaption}  
\usepackage{graphicx}    
\usepackage{xcolor}      

\usepackage{tikz}
\usepackage{pgfplots}
\usepgfplotslibrary{groupplots}

\usepackage{booktabs}
\usepackage[table]{xcolor}
\usepackage{siunitx}

\usepackage{wrapfig}   
\usepackage{booktabs}

\usepackage{pifont}

\title{ALAM: Algebraically Consistent Latent Action Model for Vision-Language-Action Models}

\author{%
  Zuojin Tang$^{1,2}$\quad
  Haoyun Liu$^{3,4,2}$\quad
  Xinyuan Chang$^{2}$\quad
  Changjie Wu$^{2}$\quad
  Dongjie Huo$^{5,2}$\\
  \textbf{Yandan Yang$^{2}$}\quad
  \textbf{Bin Liu$^{6}$}\quad
  \textbf{Zhejia Cai$^{7}$}\quad
  \textbf{Feng Xiong$^{2}$}\quad
  \textbf{Mu Xu$^{2}$}\\
  \textbf{Jiachen Luo$^{8}$}\quad
  \textbf{De Ma$^{1}$}\quad
  \textbf{Zhiheng Ma$^{4*}$}\quad
  \textbf{Gang Pan$^{1*}$}
  \\
  $^1$Zhejiang University\quad
  $^2$Amap, Alibaba Group\quad
  $^3$Nanjing University\\
  $^4$Faculty of Computility Microelectronics, Shenzhen University of Advanced Technology\\
  $^5$Beijing University of Chemical Technology\quad
  $^6$PaXini AI\\
  $^7$Tsinghua University\quad
  $^8$Queen Mary University of London\\
  \texttt{zuojintang@zju.edu.cn; mazhiheng@suat-sz.edu.cn; gpan@zju.edu.cn}\\
  $^*$Corresponding Author
}

\begin{document}

\maketitle

\begin{abstract}
Vision-language-action (VLA) models remain constrained by the scarcity of action-labeled robot data, whereas action-free videos provide abundant evidence of how the physical world changes. Latent action models offer a promising way to extract such priors from videos, but reconstruction-trained latent codes are not necessarily suitable for policy generation: they may predict future observations while lacking the structure needed to be reused or generated coherently with robot actions.
We introduce \textbf{ALAM} (\textbf{A}lgebraic \textbf{L}atent \textbf{A}ction \textbf{M}odel), an Algebraically Consistent Latent Action Model that turns temporal relations in action-free video into structural supervision. Given frame triplets, ALAM learns latent transitions that are grounded by reconstruction while being regularized by composition and reversal consistency, encouraging a locally additive transition space. For downstream VLA learning, we freeze the pretrained encoder and use its latent transition sequences as auxiliary generative targets, co-generated with robot actions under a joint flow-matching objective. This couples structured latent transitions with flow-based policy generation, allowing the policy to exploit ALAM's locally consistent transition geometry without requiring latent-to-action decoding. Representation probes show that ALAM reduces additivity and reversibility errors by 25-85$\times$ over unstructured latent-action baselines and improves long-horizon cumulative reconstruction. When transferred to VLA policies, ALAM raises the average success rate from 47.9\% to 85.0\% on MetaWorld MT50 and from 94.2\% to 98.1\% on LIBERO, with consistent gains on real-world manipulation tasks. Ablations further confirm that the strongest improvements arise from the synergy between algebraically structured latent transitions and joint flow matching. Our code is available at: \textit{https://github.com/Mark-zjtang/ALAM/tree/main}.
\end{abstract}

\begin{figure*}[t]
\vspace{-10pt}
  \begin{center}
    \centerline{\includegraphics[width=1\columnwidth]{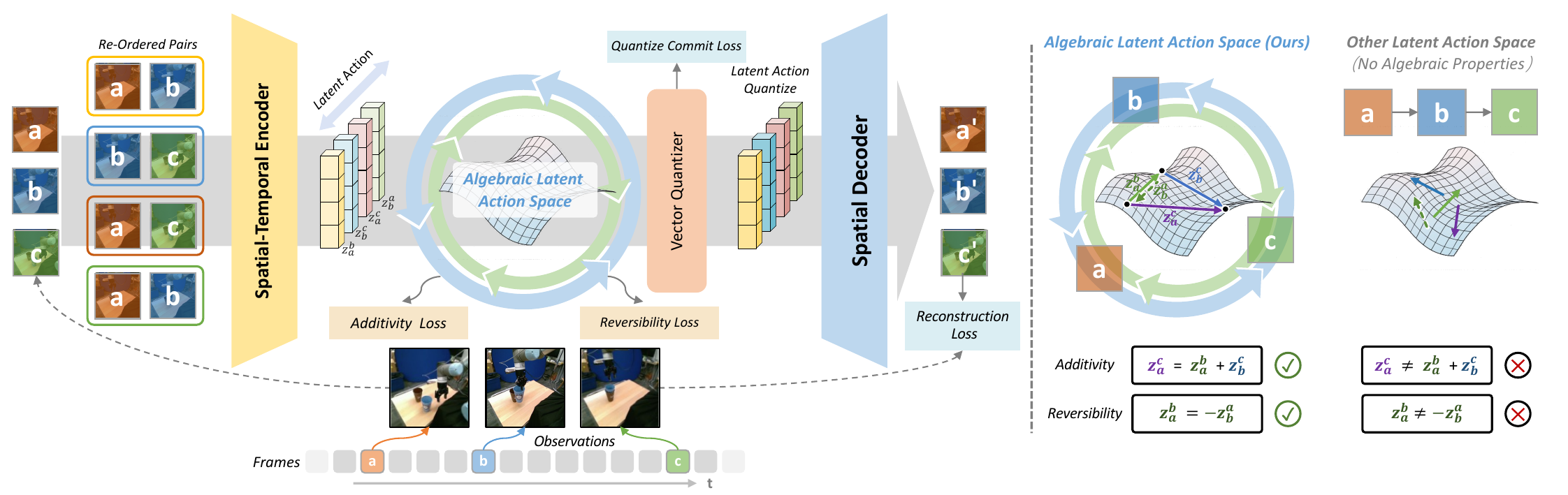}}
    \caption{\textbf{Left: Pretraining pipeline.} From a sampled frame triplet $(o_a,o_b,o_c)$, a relational encoder maps each pair $(o_i,o_j)$ to a continuous latent transition $z_i^{\,j}$, and a decoder reconstructs $\hat{o}_j$ from $o_i$ and $z_i^{\,j}$. Beyond conventional pair-based latent action models, ALAM further regularizes the latent space with additivity and reversal consistency losses. \textbf{Right: Algebraically structured latent action space.} Unlike conventional pair-based latent actions that only capture forward transitions, ALAM organizes transitions into two cycles representing forward and reverse temporal directions, with explicit additivity ($\smash{z_a^{\,c}\!\approx\!z_a^{\,b}\!+\!z_b^{\,c}}$) and sign-reversibility ($\smash{z_b^{\,a}\!\approx\!-z_a^{\,b}}$).}

    \label{ALAM}
  \end{center}
  \vspace{-26pt}
\end{figure*}

\section{Introduction}

Vision-language-action (VLA) models have emerged as a promising paradigm for robot learning, integrating visual perception, language understanding, and action generation within a unified policy. Recent systems~\cite{brohan2022rt,zitkovich2023rt,kim2024openvla,kim2025fine,chi2025diffusion,black2026pi0visionlanguageactionflowmodel,intelligence2025pi_} show that large multimodal backbones, when fine-tuned on robot demonstrations, can perform a wide range of language-conditioned manipulation tasks. However, the scalability of this paradigm remains limited by the scarcity and cost of action-labeled robot data. In contrast, action-free videos contain abundant evidence of how the physical world changes: objects are pushed, grasped, opened, stacked, and rearranged across diverse scenes and embodiments. A central challenge is therefore to extract behavior-relevant priors from such videos and transfer them effectively to downstream VLA policies.

Latent action models provide a natural route toward this goal by inferring transition variables from observation sequences without explicit action labels. Most existing methods~\cite{bruce2024genie,chen2022lapo,ye2024latent,chen2024igor,gao2025adaworld} learn such variables through reconstruction or prediction objectives, encouraging the latent code to preserve information useful for recovering future observations. While reconstruction is an important grounding signal, it does not by itself guarantee that the learned latent is useful for policy generation. A reconstruction-trained code may capture appearance changes, camera motion, background dynamics, or other predictive factors that help recover nearby frames, while still lacking the structure needed to serve as a reliable auxiliary trajectory for control. In other words, a latent representation can be predictive without being easy for a generative policy to model jointly with actions.

This limitation suggests that latent-action pretraining should not be viewed merely as learning a compact code for future-frame reconstruction. Even without action labels, videos provide more than paired examples of before-and-after observations; they also contain temporal relations among changes. A longer visual transition can be compared with the composition of shorter consecutive transitions, and a transition observed in reverse should be consistent with undoing the corresponding forward change. These relations offer a form of self-supervised structural signal that is independent of explicit robot actions. The key idea of our work is to use such temporal relations to shape the latent space, so that the learned transitions are not only predictive of visual change but also more consistent when reused over time.

We propose \textbf{ALAM} (\textbf{A}lgebraic \textbf{L}atent \textbf{A}ction \textbf{M}odel), an Algebraically Consistent Latent Action Model that learns structured latent transitions from action-free video. During pretraining, ALAM samples frame triplets from unlabeled videos and constructs both forward and reversed observation pairs. A relational encoder maps each pair to a continuous latent transition, while a decoder reconstructs the target observation from the source observation and the inferred transition. Reconstruction grounds the latent variable in visual change, whereas composition and reversal losses regularize the latent space: consecutive transitions are encouraged to agree with the longer transition they span, and a reversed transition is encouraged to cancel its forward counterpart. In this way, ALAM turns temporal relations in action-free video into structural supervision for learning reusable latent transitions.

To transfer ALAM to downstream VLA policies, we freeze the pretrained encoder and use it to extract view-specific latent transition sequences from demonstration videos, temporally aligned with robot actions. Rather than decoding these latents into actions through a separate module, we train the policy to co-generate latent transition trajectories and robot action trajectories under a shared flow-matching objective. This coupling is important: ALAM turns reconstruction-grounded transition codes into locally consistent, approximately additive auxiliary trajectories, and flow matching can exploit this structure by modeling latents and actions along compatible interpolation paths. At inference time, the latent stream is generated jointly with the action stream but is not executed directly. It instead provides an internal auxiliary trajectory that helps structure action generation over the horizon. This allows the downstream policy to exploit the locally consistent transition space learned by ALAM within the same flow-matching process used to generate robot actions.

We evaluate ALAM at three levels. At the representation level, algebraic probes show that ALAM substantially reduces additivity and reversibility errors compared with unstructured latent-action baselines, indicating that the proposed constraints reshape the latent transition space rather than merely improving reconstruction. At the prediction level, ALAM improves reconstruction fidelity at the training horizon and degrades more slowly under long-horizon cumulative composition, supporting the claim that local consistency improves temporal reuse. At the policy level, transferring the frozen encoder to VLA training yields large gains on MetaWorld MT50 and consistent improvements on LIBERO and real-world robotic tasks. Ablations further show that algebraic regularization benefits multiple generative frameworks, while the joint flow-matching formulation produces the strongest improvement, confirming the synergistic effect between structured latent transitions and flow-based policy generation. In summary, our contributions are as follows:
\begin{itemize}
    \item We identify the lack of policy-friendly structure in reconstruction-trained latent actions as a key bottleneck for transferring action-free video pretraining to VLA control.
    \item We propose ALAM, a pretraining framework that turns temporal relations in action-free video into structural supervision through composition and reversal consistency.
    \item We introduce a joint flow-matching transfer mechanism that co-generates structured latent transition trajectories and robot actions, allowing the downstream policy to exploit the structure learned during pretraining.
    \item We validate ALAM through representation-level algebraic probes, long-horizon cumulative reconstruction, and downstream VLA policy learning on simulation and real-world manipulation benchmarks.
\end{itemize}

\section{Related Work}
\subsection{Vision-Language-Action Models}

Driven by the rapid progress of Large Language Models~\cite{floridi2020gpt,achiam2023gpt,bai2023qwen,zhang2023llama,yang2025qwen3,liu2024deepseek} and Multimodal Large Language Models~\cite{wang2024qwen2,bai2025qwen3,team2023gemini,team2024gemma,team2024gemma2}, together with the emergence of large-scale robot datasets, Vision-Language-Action (VLA) models have become a dominant paradigm in robot learning. Following the RT series~\cite{brohan2022rt,zitkovich2023rt} that pioneered fine-tuning MLLMs on robot demonstrations, a line of work~\cite{kim2024openvla,kim2025fine,qu2025spatialvla,zhao2023learning,tang2025vlascd,goyal2025vla,liu2026neuralimplicitactionfields,li2024cogact} further improves scalability, controllability, and embodiment coverage. To better capture the multi-modal nature of robot actions, another line of work~\cite{chi2025diffusion,black2026pi0visionlanguageactionflowmodel,intelligence2025pi_,zheng2025x,ni2025swiftvla,shukor2025smolvla,yang2026abot} replaces deterministic decoding with diffusion or flow matching, sampling actions from noise conditioned on observations, instructions, and proprioceptive states. Since directly mapping observations to actions lacks the explicit reasoning loop of LLM-style systems, recent methods~\cite{wu2023unleashing,guo2024prediction,zhang2025up,hu2024video,cen2025worldvla,sun2026vla,liu2024deepseek,zhang2025dreamvla,cheang2024gr,ye2026world,li2026causal,xiao2025world,luo2026jointalignedlatentactionscalable} introduce an auxiliary forecasting step, either generating future frames externally or coupling pixel-level prediction with action prediction. However, such methods inherit the limitations of dense pixel reconstruction and are easily dominated by appearance details irrelevant to control. In contrast, our work bypasses pixel-level forecasting and instead learns algebraically structured latent transitions from action-free video, providing a compact, control-relevant prior for downstream policy learning.

\subsection{Latent Action Models from Videos}
Latent Action Models (LAMs) learn action-like representations directly from observation sequences to reduce the reliance on labeled robot actions. Early work~\cite{bruce2024genie,chen2022lapo,ye2024latent} shows that unlabeled videos alone can supervise latent actions via inverse-dynamics or world-model objectives. Building on this, more recent methods couple forecasting with control by asking the policy to predict compact latent futures or latent actions alongside robot actions~\cite{chen2024igor,gao2025adaworld,bu2025univla,bi2025motus,cui2024dynamo,jang2025dreamgen,zhang2025latent,lee2024behavior,kim2026hierarchical,tang2026tokenframereconsideringvisual}, while a complementary line~\cite{chen2025villa} grounds the latent space through a proprioceptive forward-dynamics module. Most of these methods do not impose explicit algebraic constraints on the latent space. A concurrent and independent preprint, AC-LAM~\cite{wei2026learning}, also studies algebraic structure in latent action spaces. While AC-LAM uses scene-wise additivity primarily to improve standalone latent-action labels, ALAM uses algebraically structured transitions as auxiliary generative trajectories within a flow-based VLA policy.

\begin{figure*}[t]
\vspace{-10pt}
  \centering
  \includegraphics[width=\textwidth]{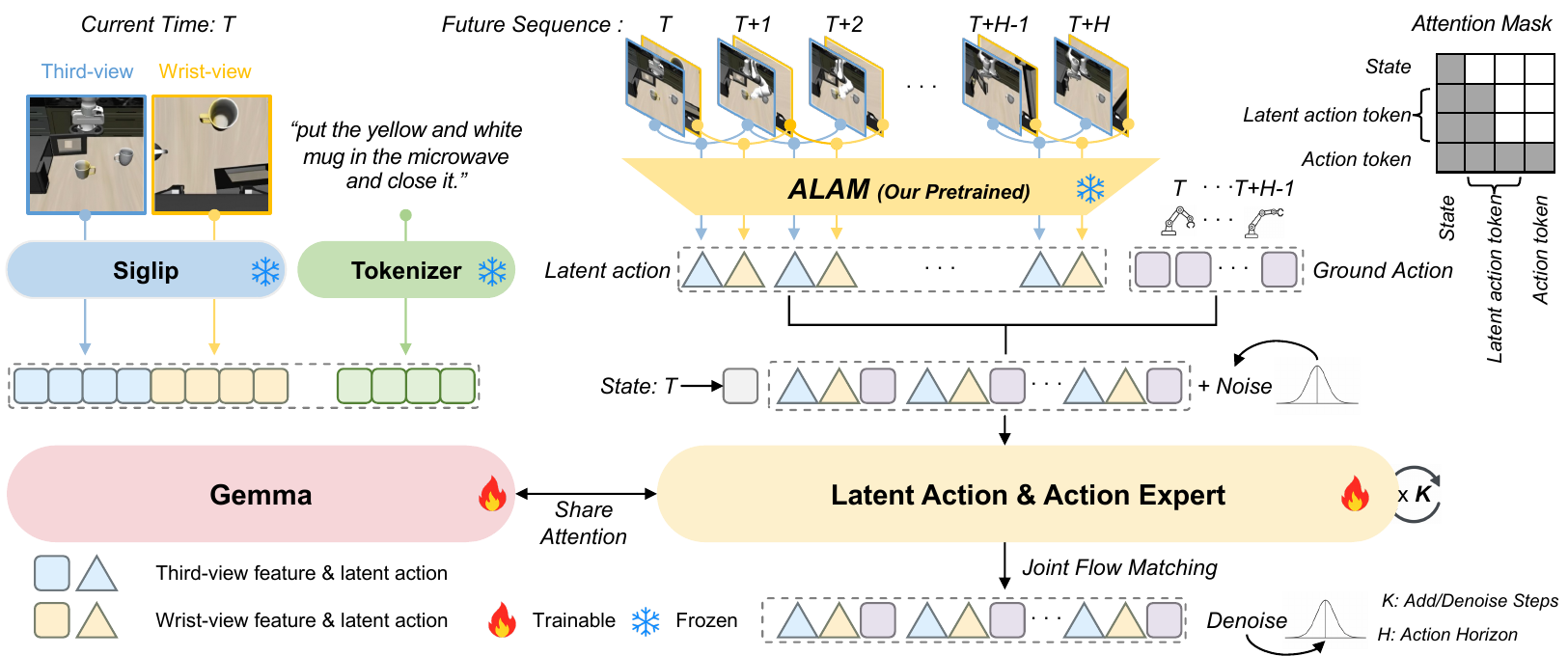}
  \caption{\textbf{Downstream transfer of ALAM.} The frozen ALAM encoder turns frame pairs over horizon $T{:}T{+}H$ into latent transition tokens, interleaved with action tokens. Conditioned on visual and language context, a shared expert co-generates both streams via $K$-step joint flow matching with the Gemma backbone; only the action stream is executed on the robot.}
  \label{fig:main_framework}
  \vspace{-8pt}
\end{figure*}

\section{Method}
We introduce \textbf{ALAM}, a two-stage framework that learns latent transitions from action-free video and transfers them to vision-language-action (VLA) policy learning. In pretraining (Sec.~\ref{sec:pretraining}), a relational encoder is trained with future-frame reconstruction together with two algebraic regularizers, composition consistency and reversal consistency, that encourage the latent transitions to be approximately additive and sign-reversible. At transfer (Sec.~\ref{sec:transfer}), the encoder is frozen and its latent transitions are jointly modeled with robot actions under a shared flow-matching objective on top of an off-the-shelf VLA backbone.
\subsection{Structured Latent Transitions}
\label{defination}
A useful representation of how a scene changes between two observations should behave like a difference: composing two consecutive transitions should match the longer one they span, and reversing a transition in time should flip its sign. For any function $f$ defined on a temporal index, the difference operator $\Delta(t_i,t_j)=f(t_j)-f(t_i)$ satisfies $\Delta(t_a,t_c)=\Delta(t_a,t_b)+\Delta(t_b,t_c)$ and $\Delta(t_b,t_a)=-\Delta(t_a,t_b)$. We use these two identities as soft constraints on a learned transition encoder. Given two observations $O_i$ and $O_j$, we represent the transition from $O_i$ to $O_j$ by
\begin{equation}
z_i^{\,j} = E(O_i, O_j),
\end{equation}
where $E(\cdot,\cdot)$ is a relational encoder that takes both endpoints as input. For three temporally ordered observations $(O_a,O_b,O_c)$ with $a<b<c$, we impose
\begin{equation}
z_a^{\,c} \approx z_a^{\,b} + z_b^{\,c}, \qquad
z_b^{\,a} \approx - z_a^{\,b},
\label{eq:transition_constraints}
\end{equation}
which we call composition consistency and reversal consistency. Both are imposed as soft regularizers, and we do not require any further structure on the latent space. See Fig.~\ref{ALAM} for the pretraining pipeline.

\subsection{Pretraining on Action-Free Video}
\label{sec:pretraining}

The pretraining stage trains a latent transition encoder on unlabeled video so that its outputs are predictive of future frames and approximately respect the two consistency relations of Sec.~\ref{defination}. Concretely, given an unlabeled video corpus $\mathcal{D}$, we draw a temporally ordered triplet and assemble four transition pairs that span both temporal directions and two horizons:
\begin{equation}
(O_a,O_b,O_c)\sim\mathcal{D},\;\; a<b<c, \qquad
\mathcal{T} \;=\; \underbrace{\{(a,b),(b,c),(a,c)\}}_{\mathcal{T}_{\mathrm{fwd}}}\;\cup\;\underbrace{\{(b,a)\}}_{\mathcal{T}_{\mathrm{rev}}} ,
\end{equation}
where $(a,b)$ and $(b,c)$ are the two short-horizon legs, $(a,c)$ provides the long-horizon target for composition consistency, and $(b,a)$ provides the time-reversed counterpart for reversal consistency.

Each pair $(O_i,O_j)\in\mathcal{T}$ is encoded into a latent transition by a relational encoder. A patch tokenizer $f_{\mathrm{patch}}$ first turns the two frames into patch sequences $X_i^{\mathrm{patch}},X_j^{\mathrm{patch}}$, which we concatenate with $K$ learnable query tokens $Q\in\mathbb{R}^{K\times d}$ along the sequence axis (denoted $[\cdot\,\|\,\cdot]$). A spatiotemporal Transformer $f_{\theta}$ processes the joint sequence; we read out the query positions, $(\cdot)^{\mathrm{query}}$, and project them through a linear head $g_{\phi}$:
\begin{equation}
z_i^{\,j} \;=\; g_{\phi}\!\Bigl( f_{\theta}\bigl([\,Q \,\|\, f_{\mathrm{patch}}(O_i,O_j)\,]\bigr)^{\mathrm{query}} \Bigr) ,
\end{equation}
yielding the four latents $\bigl[z_a^{\,b},\,z_b^{\,c},\,z_a^{\,c},\,z_b^{\,a}\bigr]$. Each latent is then passed through a learned codebook $q(\cdot)$, with $z_{q,i}^{\,j}=q(z_i^{\,j})$ and the standard commitment loss
\begin{equation}
\mathcal{L}_{\mathrm{vq}} \;=\; \bigl\|z_i^{\,j} - \operatorname{sg}[z_{q,i}^{\,j}]\bigr\|_2^2 ,
\end{equation}
where $\operatorname{sg}[\cdot]$ is the stop-gradient operator and the codebook entries are updated by EMA~\cite{van2017neural}.

For each forward pair $(i,j)\in\mathcal{T}_{\mathrm{fwd}}$, two linear heads $f_{\alpha},f_{\beta}$, a cross-attention block $h_{\psi}$, and a pixel decoder $u_{\omega}$ produce the prediction
\begin{equation}
\hat{O}^{\,i,j} \;=\; u_{\omega}\!\Bigl( h_{\psi}\bigl( [\,f_{\alpha}(z_{q,i}^{\,j}),\; f_{\beta}(X_i^{\mathrm{patch}})\,] \bigr) \Bigr) ,
\end{equation}
i.e., the decoder reconstructs $O_j$ from $z_{q,i}^{\,j}$ together with the source patches. Conditioning on $O_i$ in this way encourages the latent to carry only what is needed to map $O_i$ to $O_j$, rather than the full content of $O_j$.

The two algebraic regularizers are imposed on the continuous latents prior to quantization, so that gradients reach the encoder unimpeded by the codebook lookup:
\begin{equation}
\mathcal{L}_{\mathrm{add}} \;=\; \bigl\| z_a^{\,c} - (z_a^{\,b}+z_b^{\,c}) \bigr\|_2^2 , \qquad
\mathcal{L}_{\mathrm{rev}} \;=\; \bigl\| z_a^{\,b} + z_b^{\,a} \bigr\|_2^2 ,
\end{equation}
which are squared relaxations of the two identities in Eq.~\eqref{eq:transition_constraints}. Pixel and perceptual reconstruction terms are averaged over $\mathcal{T}_{\mathrm{fwd}}$,
\begin{equation}
\mathcal{L}_{\mathrm{rec}} = \tfrac{1}{|\mathcal{T}_{\mathrm{fwd}}|}\!\!\!\!\sum_{(i,j)\in\mathcal{T}_{\mathrm{fwd}}}\!\!\!\!\bigl\|\hat{O}^{\,i,j}-O_j\bigr\|_2^2 , \qquad
\mathcal{L}_{\mathrm{perc}} = \tfrac{1}{|\mathcal{T}_{\mathrm{fwd}}|}\!\!\!\!\sum_{(i,j)\in\mathcal{T}_{\mathrm{fwd}}}\!\!\!\!d_{\mathrm{perc}}\!\bigl(\hat{O}^{\,i,j},\,O_j\bigr) ,
\end{equation}
with $d_{\mathrm{perc}}$ the LPIPS distance~\cite{zhang2018unreasonableeffectivenessdeepfeatures}. Combining the five terms gives the total pretraining objective,
\begin{equation}
\mathcal{L}_{\mathrm{ALAM}} \;=\; \lambda_{\mathrm{vq}}\mathcal{L}_{\mathrm{vq}} \,+\, \lambda_{\mathrm{rec}}\mathcal{L}_{\mathrm{rec}} \,+\, \lambda_{\mathrm{perc}}\mathcal{L}_{\mathrm{perc}} \,+\, \lambda_{\mathrm{add}}\mathcal{L}_{\mathrm{add}} \,+\, \lambda_{\mathrm{rev}}\mathcal{L}_{\mathrm{rev}} .
\end{equation}
At transfer time, we discard the quantizer and the decoder, and freeze the encoder $E\!=\!g_{\phi}\!\circ\!f_{\theta}$ as a fixed transition extractor on robot trajectories (Sec.~\ref{sec:transfer}).

\subsection{Transfer via Joint Flow Matching}
\label{sec:transfer}

At transfer time, the goal shifts from reconstructing video frames to generating robot action sequences. We reuse the pretrained latent transitions as auxiliary generation targets in a flow-matching policy, without any fine-tuning of the encoder. As shown in Fig.~\ref{fig:main_framework}, we apply the frozen ALAM encoder to a clip of $H{+}1$ consecutive frames $V^m=(I_1^m,\dots,I_{H{+}1}^m)$ for each view $m\in\{\mathrm{th},\mathrm{wr}\}$, where $\mathrm{th}$ and $\mathrm{wr}$ denote the third-person and wrist views. Applying the encoder to every adjacent frame pair gives $H$ continuous latent transitions per view, time-aligned with the $H$ ground-truth actions $u_{1:H}$,
\begin{equation}
z_t^m \;=\; \mathcal{F}_{\mathrm{ALAM}}(I_t^m,\,I_{t+1}^m), \qquad t=1,\dots,H .
\end{equation}

We take $z_t^m$ to be the continuous output of the encoder, before the codebook lookup. The codebook in Sec.~\ref{sec:pretraining} is used only as a pretraining-time bottleneck; for joint flow matching we prefer a continuous, differentiable target with higher fidelity. Each $z_t^m$ is then linearly projected to the dimension of $u_t$, and we co-generate the two latent streams together with the action stream under a shared flow-matching objective. To let the latent transitions condition the action stream, we order the three modalities at every timestep so that the two latent tokens come first and the action token last; the full sequence concatenates these per-timestep blocks,
\begin{equation}
\bigl[\,z_1^{\mathrm{th}},\,z_1^{\mathrm{wr}},\,u_1,\;\;\dots,\;\;z_H^{\mathrm{th}},\,z_H^{\mathrm{wr}},\,u_H\,\bigr] .
\end{equation}

Let $y^m$ denote the clean sequence of modality $m\in\{u,\mathrm{th},\mathrm{wr}\}$, with $y^{\mathrm{th}}\!=\!z_{1:H}^{\mathrm{th}}$, $y^{\mathrm{wr}}\!=\!z_{1:H}^{\mathrm{wr}}$, and $y^u\!=\!u_{1:H}$. All three modalities share the same linear data-to-noise interpolation,
\begin{equation}
x_\tau^m \;=\; \tau\,\epsilon^m + (1-\tau)\,y^m, \qquad
\epsilon^m \sim \mathcal{N}(0,I), \qquad
v_{\mathrm{tgt}}^m \;=\; \epsilon^m - y^m ,
\end{equation}
sampled at a common time $\tau=0.999\,\xi+0.001$ with $\xi\sim\mathrm{Beta}(1.5,1)$~\cite{black2026pi0visionlanguageactionflowmodel}. We further apply a structured attention mask over the interleaved sequence: at each timestep, the two view-specific latent tokens attend to each other but are isolated from action tokens, while each action token attends causally to all preceding tokens. Conditioned on context $c$ (visual observations, language instructions, and proprioceptive states), the model predicts a per-modality velocity field $\hat{v}_\theta^m(x_\tau^m\mid\tau,c)$, trained with the per-modality flow-matching loss
\begin{equation}
\mathcal{L}^m \;=\; \mathbb{E}\!\left[\bigl\|\hat{v}_\theta^m(x_\tau^m\mid\tau,c)-v_{\mathrm{tgt}}^m\bigr\|_1\right] , \qquad  m\in\{u,\mathrm{th},\mathrm{wr}\}
\end{equation}
and the total transfer objective is the weighted sum
\begin{equation}
\mathcal{L}_{\mathrm{transfer}} \;=\; \lambda_{\mathrm{th}} \mathcal{L}^{\mathrm{th}} + \lambda_{\mathrm{wr}} \mathcal{L}^{\mathrm{wr}} + \lambda_u \mathcal{L}^{u} .
\end{equation}

This design connects the two stages without retraining: the latent transitions used downstream come directly from the frozen encoder, and the policy is asked to generate them jointly with the actions, along the same interpolation schedule. The action stream is therefore not predicted in isolation, but co-denoised with a structured visual signal already shaped during pretraining.

At inference time, we initialize all modalities from Gaussian noise, $x_1^m\!\sim\!\mathcal{N}(0,I)$ for $m\in\{u,\mathrm{th},\mathrm{wr}\}$, and jointly integrate from $\tau{=}1$ to $\tau{=}0$ with the learned velocity fields,
\begin{equation}
x_{\tau-\Delta_\tau}^m \;=\; x_\tau^m \,-\, \Delta_\tau\,\hat{v}_\theta^m(x_\tau^m\mid\tau,c) ,
\end{equation}
following the same interleaved token order and attention mask as in training. Both streams are denoised jointly, but only $u$ is executed on the robot.

\begin{table}[t]
\centering
\caption{\textbf{Per-difficulty success rates on MetaWorld MT50.} \textbf{Avg.} is the macro-average over the four tiers. AR: autoregressive, FM: flow matching, VA: video-augmented, LA: latent action. To our knowledge, no prior latent-action VLA reports MT50 results under this protocol.}
\label{tab:metaworld_main}
\small
\setlength{\tabcolsep}{6pt}
\renewcommand{\arraystretch}{1.1}
\begin{tabular}{l c c | c c c c | c}
\toprule
\textbf{Method} & \textbf{Type} & \textbf{Size}
& \textbf{Easy} & \textbf{Medium} & \textbf{Hard} & \textbf{Very Hard}
& \textbf{Avg.} \\
\midrule
Diffusion Policy   & DP & 0.3B & 23.1 & 10.7 & 1.9  & 6.1  & 10.5 \\
\midrule
TinyVLA-H          & AR & 1.3B & 77.6 & 21.5 & 11.4 & 15.8 & 31.6 \\
RT-2$^{*}$         & AR & 7B   & 75.5 & 35.3 & 30.7 & 15.2 & 39.2 \\
RoboTron Mani      & AR & 4B   & 85.5 & 67.7 & 76.7 & 81.0 & 77.7 \\
\midrule
GR-1$^{*}$         & VA & 0.2B & 76.6 & 35.3 & 46.0 & 44.0 & 50.5 \\
PAD                & VA & --   & 81.8 & 65.1 & 56.7 & 87.2 & 72.7 \\
Evo-1              & VA & 0.8B & 89.2 & 76.8 & 77.2 & 79.2 & 80.6 \\
\midrule
\rowcolor{gray!12}$\pi_0$ & FM & 3B & 71.8 & 48.2 & 41.7 & 30.0 & 47.9 \\
SmolVLA            & FM & 2B   & 87.1 & 51.8 & 70.0 & 64.0 & 68.2 \\
\midrule
\rowcolor{blue!8}
$\pi_0$ + ALAM \textbf{(ours)} & LA & 3B
& \textbf{89.3} & \textbf{83.6}
& \textbf{85.0} & \textbf{82.0}
& \textbf{85.0} \\
\bottomrule
\multicolumn{8}{l}{\scriptsize $^{*}$ GR-1 and RT-2$^{*}$ scores are reproduced from PAD~\cite{seo2023masked}.}
\end{tabular}
\end{table}

\begin{table}[t]
\centering
\caption{\textbf{Success rates on LIBERO.} \textbf{Avg.} is the mean over four tasks. AR: autoregressive, FM: flow matching, VA: video-augmented, LA: latent action. For latent-action methods we annotate the underlying backbone in parentheses, so improvements over the corresponding backbone (gray row) directly reflect the contribution of the latent-action design.}
\label{tab:libero_main}
\small
\setlength{\tabcolsep}{8pt}
\renewcommand{\arraystretch}{1.1}
\begin{tabular}{l c c | c c c c | c}
\toprule
\textbf{Method} & \textbf{Type} & \textbf{Size}
& \textbf{Spatial} & \textbf{Object} & \textbf{Goal} & \textbf{Long}
& \textbf{Avg.} \\
\midrule
OpenVLA              & AR & 7B   & 84.7 & 88.4 & 79.2 & 53.7 & 76.5 \\
SpatialVLA           & AR & 4B   & 88.2 & 89.9 & 78.6 & 55.5 & 78.1 \\
CoT-VLA              & AR & 7B   & 87.5 & 91.6 & 87.6 & 69.0 & 83.9 \\
\midrule
WorldVLA             & VA & 7B   & 87.6 & 96.2 & 83.4 & 60.0 & 81.8 \\
DreamVLA             & VA & 0.4B & 97.5 & 94.0 & 89.5 & 89.5 & 92.6 \\
\midrule
SmolVLA              & FM & 2B   & 93.0 & 94.0 & 91.0 & 77.0 & 88.8 \\
GR00T-N1             & FM & 2B   & 94.4 & 97.6 & 93.0 & 90.6 & 93.9 \\
\rowcolor{gray!12}$\pi_0$ & FM & 3B & 96.8 & 98.8 & 95.8 & 85.2 & 94.2 \\
$\pi_{0.5}$          & FM & 3B   & 98.8 & 98.2 & 98.0 & 92.4 & 96.9 \\
\midrule
LAPA    & LA & 7B & 87.4 & 91.2 & 90.0 & 65.4 & 83.5 \\
UniVLA  & LA & 9B & 96.5 & 96.8 & 95.6 & 92.0 & 95.2 \\
JALA    & LA & 3B & 96.0 & 98.2 & 97.4 & \textbf{96.0} & 96.9 \\
\midrule
\rowcolor{blue!8}
$\pi_0$ + ALAM \textbf{(ours)} & LA & 3B
& \textbf{99.2} & \textbf{99.6}
& \textbf{99.0} & 94.4
& \textbf{98.1} \\
\bottomrule
\end{tabular}
\end{table}

\section{Experimental Setup}
\label{sec:exp_setup}

\begin{figure*}[t]
  \centering
  \includegraphics[width=\linewidth]{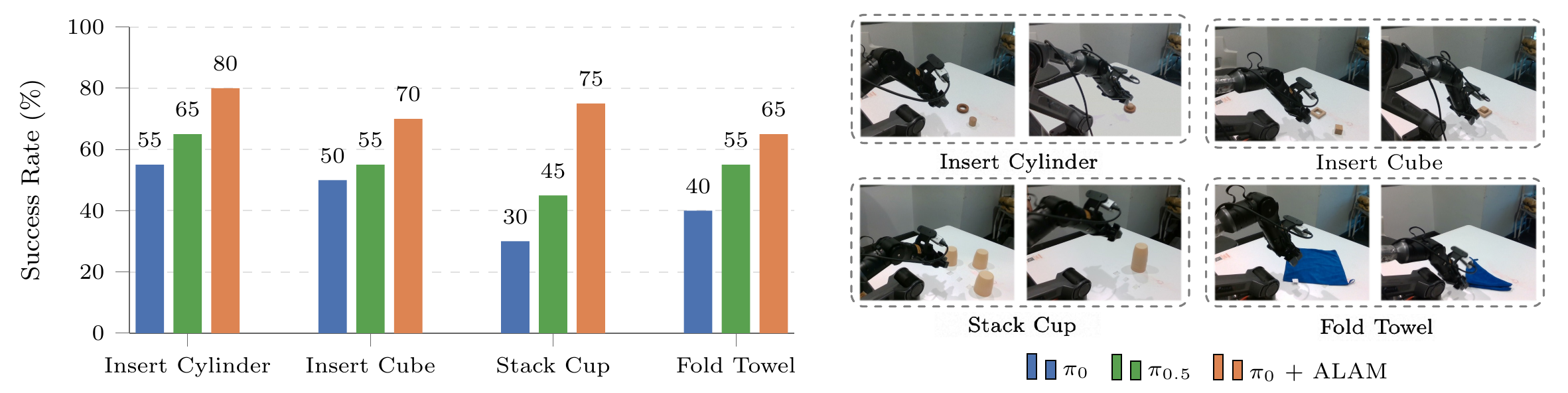}
  \caption{\textbf{Real-world results on a Piper 6-DoF manipulator.} ALAM is compared with $\pi_0$ and $\pi_{0.5}$ on four tasks: \emph{insert cylinder}, \emph{insert cube}, \emph{stack cup}, and \emph{fold towel}.}
  \label{fig:real_world_results}
\end{figure*}

\paragraph{Pretraining.} ALAM is pretrained on $11$ action-free video sources, mostly real-robot manipulation data from Open-X-Embodiment~\cite{o2024open} and CALVIN~\cite{mees2022calvin}. We use AdamW on $128\!\times\!$H20 GPUs (90G) and run for $39$ epochs (about $4$ days), at which point the pretraining loss has plateaued. The full dataset configuration and hyperparameters are in Appendix~\ref{app:impl}.

\paragraph{Downstream control.} The transfer stage is built on $\pi_0$~\cite{black2026pi0visionlanguageactionflowmodel}. The frozen ALAM encoder takes $H{+}1$ third-person and wrist frames and produces $H$ latent transitions, which are linearly projected and interleaved with the action stream (Sec.~\ref{sec:transfer}). Policies are finetuned with flow matching on PaliGemma-2B~\cite{team2024gemma} and Gemma-300M~\cite{team2024gemma} on $8\!\times\!$H20 GPUs. The full configuration is in Appendix~\ref{app:downstream}.

\paragraph{Evaluation benchmarks.} We use three settings. \textbf{(i) MetaWorld MT50}~\cite{mclean2025metaworld}: a single policy is trained on all $50$ tabletop tasks. Following Evo-1~\cite{lin2025evo}, we report per-task success rates averaged over $10$ trials per task and group tasks into Easy, Medium, Hard, and Very Hard tiers. All methods are trained for 30k steps. \textbf{(ii) LIBERO}~\cite{liu2023libero}: four suites (Spatial, Object, Goal, Long), with the Long suite focusing on long-horizon and cross-scene generalization. We report success rates over $500$ evaluation episodes per suite, again with 30k training steps. Training and inference details are in Appendix~\ref{tab:app_libero_infer}. \textbf{(iii) Real-world}: a Piper 6-DoF manipulator evaluated on four tasks (\emph{insert cylinder}, \emph{insert cube}, \emph{stack cup}, \emph{fold towel}); task descriptions and evaluation protocols are in Appendix~\ref{app:real_world}.

\paragraph{Baselines.} We use $\pi_0$~\cite{black2026pi0visionlanguageactionflowmodel} without latent actions or joint generation as the backbone reference, and compare against four families of VLA models: autoregressive (RT-2~\cite{zitkovich2023rt}, OpenVLA~\cite{kim2024openvla}), flow-matching ($\pi_{0.5}$~\cite{intelligence2025pi_}), video-augmented (Evo-1~\cite{lin2025evo}, PAD~\cite{guo2024prediction}, GR-1~\cite{wu2023unleashing}, DreamVLA~\cite{zhang2025dreamvla}, WorldVLA~\cite{cen2025worldvla}), and latent-action (UniVLA~\cite{bu2025univla}, JALA~\cite{luo2026jointalignedlatentactionscalable}, LAPA~\cite{ye2024latent}). We also include LAM~\cite{bruce2024genie} as an unstructured latent-action baseline at the same encoder capacity: it is trained on image pairs without any algebraic constraint, while ALAM uses consecutive triplets with the reversal ($\mathcal{L}_{\mathrm{rev}}$) and additivity ($\mathcal{L}_{\mathrm{add}}$) losses. Per-benchmark protocols are in Appendix~\ref{app:eval}.

\begin{figure*}[t]
\vspace{-10pt}
  \centering
  \includegraphics[width=\linewidth]{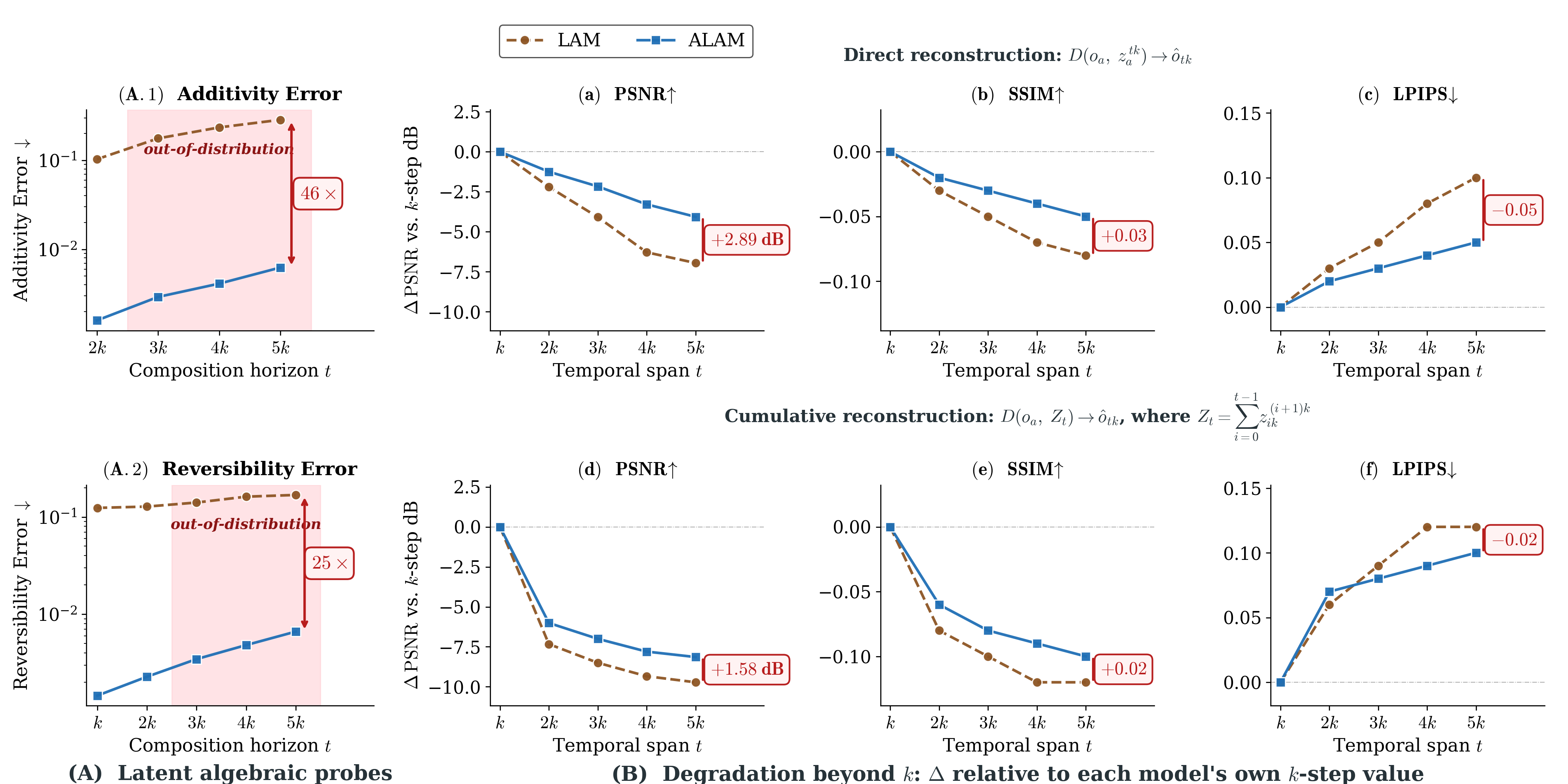}
  \caption{%
    \textbf{LAM vs.\ ALAM on the $5\%$ held-out split.}
    \textbf{(A)} Algebraic probes (log $y$, shaded $t{\geq}3k$ not seen during training):
    \textbf{(A.1)} additivity error and \textbf{(A.2)} reversibility error,
    with LAM/ALAM ratios at $t{=}5k$ shown in red.
    \textbf{(B)} Per-horizon $\Delta$ from each model's own $k$-step score,
    for direct (top, (a)--(c)) and cumulative (bottom, (d)--(f)) reconstruction.
    See Sec.~\ref{sec:probes} for definitions and $k$-step scores.
  }
  \label{fig:lam_reconstruction_and_error}
\vspace{-15pt}
\end{figure*}

\subsection{Evaluating the algebraic structure of latent actions}
\label{sec:probes}
We split the pretraining video corpus into $95\%/5\%$ train/test partitions, train ALAM and the LAM baseline on the $95\%$ split, and evaluate the frozen encoder $E$ and decoder $D$ on the held-out $5\%$. All probes use anchor and target frames at horizons $t\in\{k,2k,3k,4k,5k\}$ with stride $k{=}5$ frames; training only supervises $t\in\{k,2k\}$, so $t\in\{3k,4k,5k\}$ are not seen during pretraining. We write $z_a^{\,b}=E(o_a,o_b)$ for the latent transition between two frames, and $Z_t=\sum_{i=0}^{t-1}z_{ik}^{\,(i+1)k}$ for the sum of $t$ consecutive short-horizon latents. We look at three things: how well each model reconstructs the target frame at long horizons, how closely the latent space respects the algebraic identities, and whether the same additivity carries over to videos from a different domain.

\textbf{Reconstruction quality across horizons.} For each anchor and horizon we report PSNR, SSIM, and LPIPS between the ground-truth target $o_{tk}$ and two reconstructions: the direct one, $D(o_a,z_a^{\,tk})\!\to\!\hat{o}_{tk}$, and the cumulative one, $D(o_a,Z_t)\!\to\!\hat{o}_{tk}$. At the training horizon $t{=}k$, ALAM and LAM reach PSNR $28.87$ vs.\ $26.74$~dB, SSIM $0.91$ vs.\ $0.90$, and LPIPS $0.08$ vs.\ $0.08$. Fig.~\ref{fig:lam_reconstruction_and_error}(B) plots the per-horizon $\Delta$ relative to each model's own $k$-step score. Under both regimes ALAM keeps a smaller $\Delta$ than LAM at every $t{>}k$, and the PSNR gap at $t{=}5k$ reaches $+2.89$~dB (direct) and $+1.58$~dB (cumulative).

\textbf{Algebraic consistency on unseen horizons.} On the same latents we compute two consistency probes, the additivity error $\mathrm{Add}(t)=\mathbb{E}\,\lVert z_a^{\,tk}-Z_t\rVert$ and the reversibility error $\mathrm{Rev}(t)=\mathbb{E}\,\lVert z_a^{\,tk}+z_{tk}^{\,a}\rVert$. Both are zero for an ideally additive encoder, and $\mathrm{Add}(1)$ is zero by construction. Fig.~\ref{fig:lam_reconstruction_and_error}(A) shows that LAM's errors grow with horizon, while ALAM stays one to two orders of magnitude lower across the whole range. The largest LAM/ALAM ratio is about $85\!\times$ on reversibility at $t{=}k$, and at $t{=}5k$ the ratio is $46\!\times$ for additivity and $25\!\times$ for reversibility. The same gap is preserved on the horizons $t\in\{3k,4k,5k\}$ that were never used during training.

\textbf{Cross-domain additivity of latent actions.} We also look at this additivity qualitatively, on samples from the pretraining mixture and on game videos that the encoder never saw during training. Given an anchor $o_a$ and two endpoints $o_b,o_c$, decoding $z_a^{\,b}+z_b^{\,c}$ gives a frame that looks similar to the direct long-horizon reconstruction $D(o_a,z_a^{\,c})$, on both the test backgrounds and the unseen game footage. Additional examples are in Appendix~\ref{app:more_transfer}.

\begin{table*}[t]
\centering

\begin{minipage}[t]{0.49\textwidth}
\centering
\captionof{table}{\textbf{Generative framework ablation on MetaWorld MT50.}
\emph{LAM}/\emph{ALAM}: unstructured/structured latents. \textbf{J.}: \ding{51} joint, \ding{55} two-stage.  DP: diffusion, FM: flow matching.}
\label{tab:ablation_latent_usage}
\small
\setlength{\tabcolsep}{1.2pt}
\renewcommand{\arraystretch}{1.1}
\begin{tabular}{l c c | c c c c | c}
\toprule
\textbf{Method} & \textbf{J.} & \textbf{Gen.}
& \textbf{Easy} & \textbf{Med.} & \textbf{Hard} & \textbf{V.H.}
& \textbf{Avg.} \\
\midrule
\rowcolor{gray!12}
$\pi_0$  & --        & FM & 71.8 & 48.2 & 41.7 & 30.0 & 47.9 \\
\midrule
+ LAM    & \ding{51} & DP & 70.0 & 37.3 & 20.0 & 26.0 & 38.3 \\
+ ALAM   & \ding{51} & DP & 83.2 & 63.6 & 41.7 & 74.0 & 65.6 \\
\midrule
+ LAM    & \ding{51} & FM & 82.5 & 44.5 & 55.0 & 78.0 & 65.0 \\
+ ALAM   & \ding{55} & FM & 83.2 & 53.6 & 56.7 & 82.0 & 68.9 \\
\rowcolor{blue!8}
+ ALAM \textbf{(ours)} & \ding{51} & FM
& \textbf{89.3} & \textbf{83.6}
& \textbf{85.0} & \textbf{82.0}
& \textbf{85.0} \\
\bottomrule
\end{tabular}
\end{minipage}
\hfill
\begin{minipage}[t]{0.49\textwidth}
\centering

\captionof{table}{\textbf{Loss ablation on MetaWorld MT50.}
$\mathcal{L}_{\mathrm{rev}}$ and $\mathcal{L}_{\mathrm{add}}$ are added to the reconstruction loss to form the full ALAM objective. All variants use the epoch-$39$ checkpoint.}
\label{tab:loss_ablation}
\small
\setlength{\tabcolsep}{3pt}
\renewcommand{\arraystretch}{1.1}
\begin{tabular}{l | c c c c | c}
\toprule
\textbf{Variant} & \textbf{Easy} & \textbf{Med.} & \textbf{Hard} & \textbf{V.H.} & \textbf{Avg.} \\
\midrule
\rowcolor{gray!12}$\pi_0$
& 71.8 & 48.2 & 41.7 & 30.0 & 47.9 \\
\midrule
w/o VQ
& 84.6 & 71.8 & 66.7 & 60.0 & 70.8\\
\midrule
w/o $\mathcal{L}_{\mathrm{rev}}{+}\mathcal{L}_{\mathrm{add}}$
& 82.9 & 60.0 & 51.7 & 88.0 & 70.7\\
w/o $\mathcal{L}_{\mathrm{rev}}$
& 87.1 & 60.9 & 48.3 & \textbf{90.0} & 71.6 \\
w/o $\mathcal{L}_{\mathrm{add}}$
& 84.6 & 60.0 & 61.7 & \textbf{90.0} & 74.1\\
\rowcolor{blue!8}
ALAM \textbf{(ours)}
& \textbf{89.3} & \textbf{83.6}
& \textbf{85.0} & 82.0
& \textbf{85.0} \\
\bottomrule
\end{tabular}
\end{minipage}

\end{table*}

\subsection{Downstream Policy Learning}
\label{sec:exp_policy}
\textbf{Main results.} We now turn to downstream control and evaluate whether the pretrained structured transitions translate into better policies. On MetaWorld MT50, adding ALAM latent transitions to the $\pi_0$ backbone raises the average success rate from $47.9\%$ to $\mathbf{85.0\%}$, a $+37.1$-point absolute gain, and also exceeds the strongest video-augmented baseline Evo-1~\cite{lin2025evo} by $4.4$ points on the same protocol. On LIBERO, ALAM lifts $\pi_0$ from $94.2\%$ to $\mathbf{98.1\%}$ averaged across the four suites, with the largest relative gain on the long-horizon LIBERO-Long split, the suite most sensitive to compositional reasoning. On the real-world Piper robot (Fig.~\ref{fig:real_world_results}), ALAM improves over $\pi_0$ and $\pi_{0.5}$~\cite{intelligence2025pi_} on all four tasks, with the largest gain reaching 45 percentage points over $\pi_0$ on the \textit{Stack Cup} task. The three settings together indicate that the pretrained transitions transfer from action-free video to physical manipulation, rather than fitting a particular simulator.

\textbf{ALAM and flow matching combine well.}
Tab.~\ref{tab:ablation_latent_usage} varies the latent structure (LAM vs.\ ALAM) and the policy coupling (two-stage vs.\ joint), under both diffusion and flow matching, on MetaWorld~MT50. Three patterns stand out. (i)~Adding LAM with joint diffusion reaches $38.3\%$, below $\pi_0$ alone ($47.9\%$), so joint coupling does not help when the latent stream is unstructured. (ii)~Replacing LAM with ALAM at the same coupling raises the average from $38.3\%$ to $65.6\%$ under diffusion and from $65.0\%$ to $\mathbf{85.0\%}$ under flow matching, the largest single change in either column.
(iii)~With ALAM fixed under flow matching, switching from two-stage to joint coupling raises the average from $68.9\%$ to $\mathbf{85.0\%}$ ($+16.1$~pts), and the joint flow-matching variant is also the best column overall. The pattern is consistent with the additivity and reversibility errors reported in Sec.~\ref{sec:probes}.

\textbf{Ablation on the algebraic losses.}
A natural confound when comparing ALAM with LAM is that ALAM additionally consumes triplet inputs at pretraining, so its gain over LAM could in principle reflect added input capacity rather than the algebraic constraints. To control for this, we train five encoder variants on the same single-domain pretraining subset and transfer each to $\pi_0$ with joint flow matching under matched downstream conditions (Tab.~\ref{tab:loss_ablation}; epoch-$39$ checkpoints). Removing both algebraic losses (\emph{w/o $\mathcal{L}_{\mathrm{rev}}{+}\mathcal{L}_{\mathrm{add}}$}) reaches $70.7\%$, only $+5.7$~pts over the LAM baseline ($65.0\%$) and well below the full ALAM at $\mathbf{85.0\%}$ ($+20.0$~pts over LAM). Each loss on its own recovers most of this gap (\emph{w/o $\mathcal{L}_{\mathrm{rev}}$}: $71.6\%$; \emph{w/o $\mathcal{L}_{\mathrm{add}}$}: $74.1\%$), and using both gives the highest score. Within this controlled comparison, the structured latent transitions, rather than the additional inputs, account for most of the downstream gain.

\section{Conclusion and Discussion}
\label{sec:conclusion_and_discussion}

We presented ALAM, a pretraining framework that learns structured
latent transitions from action-free video by regularizing a relational
encoder with two algebraic constraints, composition consistency and
reversal consistency, motivated by elementary identities of difference
operators.
The resulting latents are trained to be approximately additive and
sign-reversible, and transfer best when the latent is co-generated
with the robot action under a shared flow-matching objective rather
than decoded through a separate module.
On a held-out video split, ALAM reduces the additivity and
reversibility errors by up to $25$--$85\!\times$ relative to an
unstructured latent-action baseline, and the cumulative reconstruction
quality degrades more slowly with horizon.
Transferring ALAM into a $\pi_0$ backbone matches or exceeds the best
previously reported numbers we are aware of on MetaWorld~MT50, with
positive transfer on LIBERO and on a four-task real-world Piper setup.
Composition and reversal consistency each improve over an unstructured
LAM under both diffusion and flow matching, and the joint
flow-matching variant gives the largest improvement in our ablations.

\paragraph{Limitations.}
ALAM provides a local and approximate algebraic regularizer rather than an exact group-structured latent space. Although the learned transitions reduce composition and reversal residuals and generalize to held-out temporal spans, the triplet-based constraints do not guarantee exact consistency or extrapolation to arbitrary horizons. In addition, joint flow matching introduces auxiliary latent tokens at inference time, increasing computation relative to action-only generation. However, these tokens are not decoded into future frames and are not executed by the robot, so the additional cost is confined to the denoising network.

\paragraph{Acknowledgements.} 
This work is supported by the National Natural Science Foundation of China under Grant 62576224; the Guangdong Provincial Key Laboratory of Computility Microelectronics (2024B1212010007); the Guangdong Basic and Applied Basic Research Foundation Project (2026A1515011733); the Characteristic Innovation Project of Ordinary Universities in Guangdong Province (2024KTSCX026); the Research and Development Fund Project of China Huaneng Group Co., Ltd. (HNKJ24-H144).

\bibliography{neurips_2026}
\bibliographystyle{plain}

\clearpage
\appendix

\section{Implementation Details}
\label{app:impl}

\subsection{ALAM pretraining}
\label{app:pretrain_details}

The relational encoder $E$ maps a frame pair $(O_i, O_j)$ to a latent transition $z_i^{\,j}\in\mathbb{R}^{7\times128}$. Frames are tokenized with a $14\!\times\!14$ patch embedding, and the patch sequence is prepended with 7 learnable latent queries and processed by a $12$-layer ViT. The query outputs are linearly projected to the latent transition space. During pretraining, $z_i^{\,j}$ passes through a small VQ bottleneck and is fed to a symmetric $12$-layer decoder that reconstructs $\hat O_j$ from $O_i$ and the quantized latent. Architectural hyperparameters are listed in Tab.~\ref{tab:app_arch}. 

In the implementation summarized in Tab.~\ref{tab:app_arch}, seven learnable transition-query tokens serve as action-query slots, producing a continuous transition tensor of shape [B, 7, 128], where B denotes the batch size. Separately, VectorQuantizer2 is configured with 256 shared codebook entries, resulting in an embedding table of shape [256, 128]. Here, the latent dimension of 128 specifies the feature dimension of each transition token and each codebook entry. The quantizer independently maps each of the seven tokens to one of these 256 shared entries and returns indices of shape [B, 7]. There is no intermediate pooling from 256 encoder queries to seven transition tokens.

\begin{table}[h]
\centering
\small
\setlength{\tabcolsep}{8pt}
\caption{ALAM pretraining architecture.}
\label{tab:app_arch}
\begin{tabular}{ll}
\toprule
\textbf{Component} & \textbf{Value} \\
\midrule
Input resolution             & $200\!\times\!200\!\times\!3$ (RGB) \\
Patch size                   & $14$ \\
Hidden size / heads          & $768$ / $12$ \\
Encoder / decoder depth      & $12$ / $12$ \\
\midrule
Shared VQ entries               & $256$ \\
Latent dimension             & $128$ \\
Learnable transition-query tokens             & $7$ \\
\bottomrule
\end{tabular}
\end{table}

The pretraining loss combines pixel reconstruction, an LPIPS perceptual term~\cite{zhang2018unreasonableeffectivenessdeepfeatures}, the two algebraic regularizers (composition and reversal consistency), and the standard VQ commitment loss. All terms are weighted equally at $1.0$ (Tab.~\ref{tab:app_loss_weights}). For each video, we first apply temporal subsampling with a frame-skip of $5$ by default and $10$ for CALVIN. We then sample a temporally ordered triplet $(O_a, O_b, O_c)$ from the subsampled
sequence, with $a < b < c$. The four pairs $\{(a,b),(b,c),(a,c),(b,a)\}$ are encoded independently and supervised as in Sec.~\ref{sec:pretraining}. We optimize with AdamW (lr $10^{-4}$, weight decay $10^{-4}$, $\beta_1{=}0.9$, $\beta_2{=}0.95$) and per-GPU batch size $32$ for $39$ epochs, on $128\!\times\!$H20 ($90$\,GB) GPUs (about $4$--$5$ days).

\begin{table}[h]
\centering
\small
\setlength{\tabcolsep}{10pt}
\caption{Pretraining loss weights.}
\label{tab:app_loss_weights}
\begin{tabular}{lcl}
\toprule
\textbf{Term} & \textbf{Weight} & \textbf{Role} \\
\midrule
$\lambda_{\mathrm{rec}}$    & $1.0$ & Pixel reconstruction of $\hat O_j$ \\
$\lambda_{\mathrm{perc}}$   & $1.0$ & LPIPS perceptual similarity \\
$\lambda_{\mathrm{add}}$    & $1.0$ & Composition consistency \\
$\lambda_{\mathrm{rev}}$    & $1.0$ & Reversal consistency \\
$\lambda_{\mathrm{commit}}$ & $1.0$ & VQ commitment \\
\bottomrule
\end{tabular}
\end{table}

The $11$ video sources from Sec.~\ref{sec:exp_setup} are mixed with the sampling weights in Tab.~\ref{tab:app_data_mix}, and clips are drawn with probability proportional to their weight. Frames are decoded at the corresponding stride and resized to $200\!\times\!200$. No action, language, or proprioceptive labels are used at this stage.

\begin{table}[h]
\centering
\small
\setlength{\tabcolsep}{7pt}
\caption{Pretraining data mixture and sampling weights. Stride is the frame-skip used for decoding.}
\label{tab:app_data_mix}
\begin{tabular}{lcc}
\toprule
\textbf{Dataset} & \textbf{Stride} & \textbf{Weight} \\
\midrule
fractal20220817 (RT-1)         & $5$  & $150$ \\
BridgeData-V2                  & $5$  & $50$  \\
TACO-Play                      & $5$  & $5$   \\
JaCo-Play                      & $5$  & $20$  \\
Berkeley Cable Routing         & $5$  & $20$  \\
RoboTurk                       & $5$  & $10$  \\
NYU Door-Opening               & $5$  & $5$   \\
VIOLA                          & $5$  & $3$   \\
Berkeley AutoLab UR5           & $5$  & $5$   \\
TOTO                           & $5$  & $5$   \\
CALVIN (\texttt{task\_ABC\_D}) & $10$ & $200$ \\
\bottomrule
\end{tabular}
\end{table}

\subsection{Downstream adaptation}
\label{app:downstream}

The transfer stage is built on $\pi_0$~\cite{black2026pi0visionlanguageactionflowmodel}, with PaliGemma-2B~\cite{team2024gemma} as the vision-language encoder and Gemma-300M as the action expert. We do not add a latent-to-action decoder; latent transitions enter the policy through the input pathway in Sec.~\ref{sec:transfer}. For each view, the frozen ALAM encoder turns $H{+}1$ consecutive frames into $H$ continuous latent transitions, which are linearly projected to the action dimension and interleaved with the action stream. The encoder $E$ is frozen throughout, and only the $\pi_0$ backbone, action expert, and the linear projection are updated. Policies are trained with flow matching on $8\!\times\!$H20 GPUs; default hyperparameters are in Tab.~\ref{tab:app_downstream_hp}, with per-task overrides (e.g., training steps for short vs.\ long horizons) noted in the corresponding benchmark protocol.

\begin{table}[h]
\centering
\small
\setlength{\tabcolsep}{10pt}
\caption{Downstream finetuning hyperparameters (defaults).}
\label{tab:app_downstream_hp}
\begin{tabular}{ll}
\toprule
\textbf{Hyperparameter} & \textbf{Value} \\
\midrule
Optimizer            & AdamW ($\beta_1{=}0.9$, $\beta_2{=}0.95$) \\
Learning rate        & $5\!\times\!10^{-5}$ \\
Weight decay         & $10^{-4}$ \\
Per-GPU batch size   & $32$ \\
GPUs                 & $8\!\times\!$H20 ($90$\,GB) \\
Action chunk $H$     & $16$ \\
Vision resolution    & $224\!\times\!224$ \\
Frozen modules       & ALAM encoder $E$ \\
Trainable modules    & PaliGemma-2B, Gemma-300M, linear projection \\
\bottomrule
\end{tabular}
\end{table}

\section{Simulation Benchmark Protocols}
\label{app:eval}

On MetaWorld~MT50~\cite{mclean2025metaworld}, we train a single multi-task policy on all $50$ tasks and follow the Evo-1~\cite{lin2025evo} protocol: each task is evaluated with $10$ rollouts of up to $200$ environment steps, with success defined by the task-specific success flag. We report per-tier means (Easy / Medium / Hard / Very Hard, with the same partitioning as Evo-1) and the overall average; the full task split is in App.~\ref{app:task_definitions}. The action horizon is $5$ at both training and inference for $\pi_0$ and $\pi_0$+ALAM.

On LIBERO~\cite{liu2023libero}, we use the four suites (Spatial, Object, Goal, Long) and follow the original protocol: each suite is evaluated with $500$ episodes from unseen initial states, and we report per-suite success and the overall average. All four suites share a single trained checkpoint with action horizon $H{=}20$, and only the inference-time action horizon and replan interval differ across suites (Tab.~\ref{tab:app_libero_infer}).

\begin{table}[h]
\centering
\small
\setlength{\tabcolsep}{10pt}
\caption{LIBERO inference configuration. All suites share the same trained checkpoint (training $H{=}20$); only the inference action horizon and replan step differ.}
\label{tab:app_libero_infer}
\begin{tabular}{lcc}
\toprule
\textbf{Suite} & \textbf{Inference $H$} & \textbf{Replan step} \\
\midrule
Spatial & $14$ & $5$ \\
Object  & $14$ & $10$ \\
Goal    & $18$ & $7$  \\
Long    & $18$ & $12$ \\
\bottomrule
\end{tabular}
\end{table}

\section{Task Difficulty Definitions in MetaWorld MT50}
\label{app:task_definitions}

We follow~\cite{seo2023masked} and partition the MetaWorld MT50 tasks into four difficulty tiers (Tab.~\ref{tab:task_difficulty}).

\begin{table}[h]
\centering
\caption{Task difficulty partitioning for MetaWorld MT50.}
\label{tab:task_difficulty}
\footnotesize
\begin{tabular}{p{2.2cm} p{10.5cm}}
\toprule
\textbf{Difficulty} & \textbf{Task Names} \\
\midrule
\textbf{Easy (28 tasks)} & \texttt{button-press}, \texttt{button-press-topdown}, \texttt{button-press-topdown-wall}, \texttt{button-press-wall}, \texttt{coffee-button}, \texttt{dial-turn}, \texttt{door-close}, \texttt{door-lock}, \texttt{door-open}, \texttt{door-unlock}, \texttt{drawer-close}, \texttt{drawer-open}, \texttt{faucet-close}, \texttt{faucet-open}, \texttt{handle-press}, \texttt{handle-press-side}, \texttt{handle-pull}, \texttt{handle-pull-side}, \texttt{lever-pull}, \texttt{plate-slide}, \texttt{plate-slide-back}, \texttt{plate-slide-back-side}, \texttt{plate-slide-side}, \texttt{reach}, \texttt{reach-wall}, \texttt{window-close}, \texttt{window-open}, \texttt{peg-unplug-side} \\
\midrule
\textbf{Medium (11 tasks)} & \texttt{basketball}, \texttt{bin-picking}, \texttt{box-close}, \texttt{coffee-pull}, \texttt{coffee-push}, \texttt{hammer}, \texttt{peg-insert-side}, \texttt{push-wall}, \texttt{soccer}, \texttt{sweep}, \texttt{sweep-into} \\
\midrule
\textbf{Hard (6 tasks)} & \texttt{assembly}, \texttt{hand-insert}, \texttt{pick-out-of-hole}, \texttt{pick-place}, \texttt{push}, \texttt{push-back} \\
\midrule
\textbf{Very Hard (5 tasks)} & \texttt{shelf-place}, \texttt{disassemble}, \texttt{stick-pull}, \texttt{stick-push}, \texttt{pick-place-wall} \\
\bottomrule
\end{tabular}
\end{table}

\section{Real-World Tasks: Setup and Protocol}
\label{app:real_world}

\begin{figure*}[h]
  \centering
  \includegraphics[width=\textwidth]{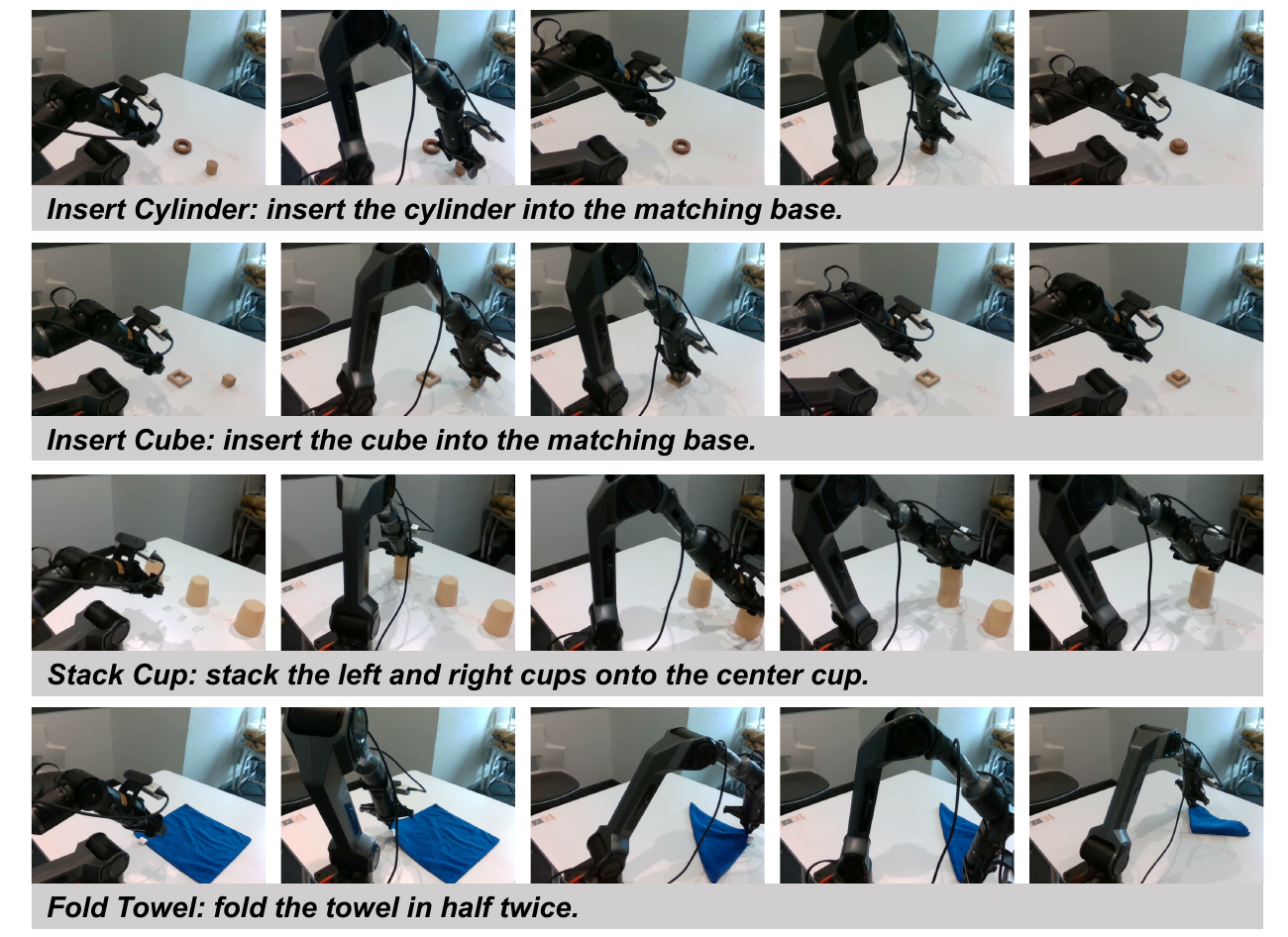}
  \caption{\textbf{Key frames of real-world rollouts from ALAM.}}
  \label{fig:real_demo}
\end{figure*}

All real-world experiments use an Agile-X Piper $6$-DoF arm with a parallel-jaw gripper. The robot is observed by two synchronized RGB cameras: a third-person Intel RealSense D435 on a tripod and a wrist-mounted RealSense D405. Both streams are resized to $224\!\times\!224$ and synchronized at the control rate. We collect $50$ teleoperated demonstrations per task ($200$ in total), used only for downstream finetuning; no real-world data is used during ALAM pretraining.

We evaluate on four tasks. \textbf{Insert Cylinder.} The robot grasps a cylinder and inserts it into a matching socket base, with one object and its base on the tabletop per trial; we run $20$ trials, and a trial counts as successful if the cylinder is correctly inserted and remains stably seated after release. The success rate is reported on a $0$--$100$ scale, equivalent to $5$ points per successful trial. \textbf{Insert Cube.} Same setup as \emph{Insert Cylinder}, but with a cube; the geometry requires more accurate orientation alignment, making the task more sensitive to execution precision. \textbf{Stack Cup.} Three cups are placed upside down on the table, and the robot stacks the left and right cups onto the center cup in sequence. We run $10$ trials; in each trial, stacking one cup gives $5$ points and both cups give $10$, so the total over $10$ trials is on a $0$--$100$ scale. \textbf{Fold Towel.} Starting from a flat towel, the robot folds it in half twice. We run $10$ trials; each successful fold gives $5$ points, so completing both folds gives $10$ per trial, and the sum across trials is on a $0$--$100$ scale.

\section{More Transfer Results}
\label{app:more_transfer}

\begin{figure*}[h]
  \centering
  \includegraphics[width=\textwidth]{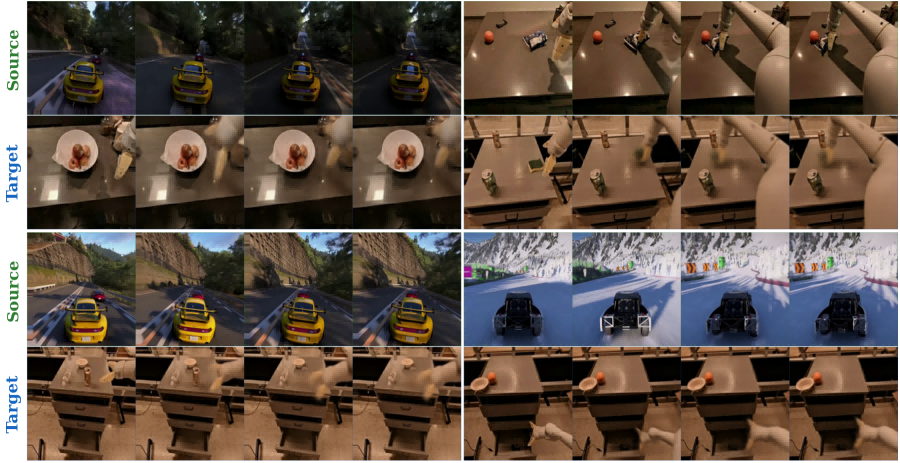}
  \caption{\textbf{Cross-domain additivity (ground-truth target reference).} Each panel has two rows. \textbf{Source} is a trajectory in the source domain; \textbf{Target} is the reconstruction in a different target domain obtained by transferring the inferred latent transitions. Columns: anchor $o_a$; forward state $\hat o_a^{\,b}$; forward state $\hat o_a^{\,c}$; and additive composition $\hat o_a^{\,b}\!+\!\hat o_b^{\,c}$, which should align with $\hat o_a^{\,c}$ if the latent transitions are additive. Hatted symbols denote reconstructions; un-hatted denote ground truth.}
  \label{fig:ALAM_transfer_gt}
\end{figure*}

\begin{figure*}[h]
  \centering
  \includegraphics[width=\textwidth]{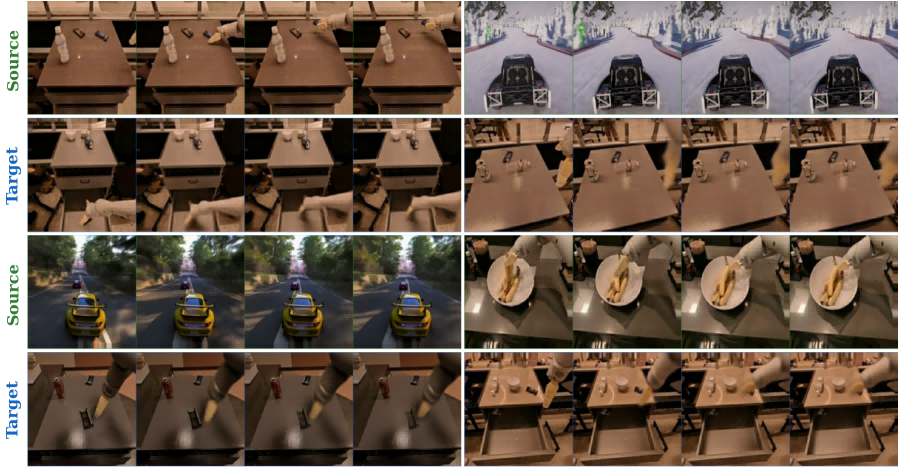}
  \caption{\textbf{Cross-domain additivity (transferred target).} Same layout as Fig.~\ref{fig:ALAM_transfer_gt}, but the target rows are produced solely from transferred latent transitions, with no ground-truth target frames as reference.}
  \label{fig:ALAM_transfer_tf}
\end{figure*}

We pick two domains from the pretraining mixture and use one as source and the other as target. Fig.~\ref{fig:ALAM_transfer_gt} uses ground-truth target-domain frames as the rendering reference, which gives an upper bound on visual fidelity that depends only on the decoder. Fig.~\ref{fig:ALAM_transfer_tf} uses transferred latent transitions as the only source of target-domain dynamics, with no ground-truth target frames as reference. In both settings, the composed reconstruction $\hat o_a^{\,b}\!+\!\hat o_b^{\,c}$ is close to the direct $\hat o_a^{\,c}$ rollout across the four source/target pairs, including pairs that differ in viewpoint, embodiment, and scene appearance. We do not claim domain-invariance in general; the qualitative match supports using the latent transitions as an auxiliary signal for downstream policies (Sec.~\ref{sec:transfer}).

\section{Inference-time intervention on the latent stream}
\label{app:latent_intervention}

As an additional check, we apply three test-time interventions to the frozen $\pi_0$+ALAM model (no retraining or finetuning). \emph{Freeze-at-noise} keeps the latent input fixed at the initial Gaussian noise sample throughout the flow-matching trajectory. \emph{Block attention} masks the action-to-latent cross-attention. \emph{Temporal shuffle} permutes the temporal order of the generated latents within each rollout while keeping their values. On MetaWorld~MT50, all three interventions lower the average success rate from $85.0\%$ to between $75$ and $77$ points (Tab.~\ref{tab:latent_intervention}); we include this as a sanity check that the latent stream is used at inference, and do not draw further conclusions from the relative ordering of the three interventions.

\begin{table}[h]
\centering
\setlength{\tabcolsep}{4pt}
\renewcommand{\arraystretch}{1.1}
\caption{Test-time interventions on the latent stream (MetaWorld~MT50, per-tier success rate \%). All rows share the same trained weights.}
\label{tab:latent_intervention}
\small
\begin{tabular}{l | c c c c | c}
\toprule
\textbf{Variant} & \textbf{Easy} & \textbf{Med.} & \textbf{Hard} & \textbf{V.Hard} & \textbf{Avg.} \\
\midrule
\rowcolor{gray!12}$\pi_0$ (no-latent baseline)
                                & 71.8 & 48.2 & 41.7 & 30.0 & 47.9 \\
\rowcolor{blue!8}
ALAM (default)                  & \textbf{89.3} & \textbf{83.6}
                                & \textbf{85.0} & \textbf{82.0}
                                & \textbf{85.0} \\
\midrule
\,\,Freeze latent               & 86.4 & 62.7 & 65.0 & 90.0 & 76.0 \\
\,\,Block A$\to$L attention     & 83.9 & 68.2 & 63.3 & 90.0 & 76.4 \\
\,\,Shuffle latent              & 84.3 & 65.5 & 63.3 & 88.0 & 75.3 \\
\bottomrule
\end{tabular}
\end{table}
\section{Broader Impacts}
\label{app:broader_impact}

This work may help reduce reliance on expensive action-labeled robot data and improve the data efficiency of robot policy learning, which can make robot learning research more accessible. As with other learned policies, real-world deployment still requires task-specific evaluation and system-level safety measures.

\clearpage
\section*{NeurIPS Paper Checklist}

\begin{enumerate}

\item {\bf Claims}
    \item[] Question: Do the main claims made in the abstract and introduction accurately reflect the paper's contributions and scope?
    \item[] Answer: \answerYes{} 
    \item[] Justification: The abstract and introduction state the main claims of the paper: ALAM learns algebraically structured latent transitions from action-free video, improves algebraic consistency and long-horizon reconstruction, and improves downstream VLA policy performance on MetaWorld, LIBERO, and real-world tasks. These claims are supported by the experimental results in Secs.~\ref{sec:probes} and~\ref{sec:exp_policy}, Tabs.~\ref{tab:metaworld_main}, \ref{tab:libero_main}, \ref{tab:ablation_latent_usage}, and \ref{tab:loss_ablation}, Fig.~\ref{fig:real_world_results}, and Appendix~\ref{app:latent_intervention}.
    \item[] Guidelines:
    \begin{itemize}
        \item The answer \answerNA{} means that the abstract and introduction do not include the claims made in the paper.
        \item The abstract and/or introduction should clearly state the claims made, including the contributions made in the paper and important assumptions and limitations. A \answerNo{} or \answerNA{} answer to this question will not be perceived well by the reviewers. 
        \item The claims made should match theoretical and experimental results, and reflect how much the results can be expected to generalize to other settings. 
        \item It is fine to include aspirational goals as motivation as long as it is clear that these goals are not attained by the paper. 
    \end{itemize}

\item {\bf Limitations}
    \item[] Question: Does the paper discuss the limitations of the work performed by the authors?
    \item[] Answer: \answerYes{} 
    \item[] Justification: The paper explicitly discusses limitations in Sec.~\ref{sec:conclusion_and_discussion}, including non-zero algebraic residuals, bounded temporal supervision, limited real-world evaluation on four tasks and one embodiment, and the added inference cost from latent-token generation.
    \item[] Guidelines:
    \begin{itemize}
        \item The answer \answerNA{} means that the paper has no limitation while the answer \answerNo{} means that the paper has limitations, but those are not discussed in the paper. 
        \item The authors are encouraged to create a separate ``Limitations'' section in their paper.
        \item The paper should point out any strong assumptions and how robust the results are to violations of these assumptions (e.g., independence assumptions, noiseless settings, model well-specification, asymptotic approximations only holding locally). The authors should reflect on how these assumptions might be violated in practice and what the implications would be.
        \item The authors should reflect on the scope of the claims made, e.g., if the approach was only tested on a few datasets or with a few runs. In general, empirical results often depend on implicit assumptions, which should be articulated.
        \item The authors should reflect on the factors that influence the performance of the approach. For example, a facial recognition algorithm may perform poorly when image resolution is low or images are taken in low lighting. Or a speech-to-text system might not be used reliably to provide closed captions for online lectures because it fails to handle technical jargon.
        \item The authors should discuss the computational efficiency of the proposed algorithms and how they scale with dataset size.
        \item If applicable, the authors should discuss possible limitations of their approach to address problems of privacy and fairness.
        \item While the authors might fear that complete honesty about limitations might be used by reviewers as grounds for rejection, a worse outcome might be that reviewers discover limitations that aren't acknowledged in the paper. The authors should use their best judgment and recognize that individual actions in favor of transparency play an important role in developing norms that preserve the integrity of the community. Reviewers will be specifically instructed to not penalize honesty concerning limitations.
    \end{itemize}

\item {\bf Theory assumptions and proofs}
    \item[] Question: For each theoretical result, does the paper provide the full set of assumptions and a complete (and correct) proof?
    \item[] Answer: \answerNA{} 
    \item[] Justification: The paper does not present formal theoretical results such as theorems, lemmas, or proofs. The algebraic relations are used as modeling principles and training regularizers rather than as formal claims requiring proof.
    \item[] Guidelines:
    \begin{itemize}
        \item The answer \answerNA{} means that the paper does not include theoretical results. 
        \item All the theorems, formulas, and proofs in the paper should be numbered and cross-referenced.
        \item All assumptions should be clearly stated or referenced in the statement of any theorems.
        \item The proofs can either appear in the main paper or the supplemental material, but if they appear in the supplemental material, the authors are encouraged to provide a short proof sketch to provide intuition. 
        \item Inversely, any informal proof provided in the core of the paper should be complemented by formal proofs provided in appendix or supplemental material.
        \item Theorems and Lemmas that the proof relies upon should be properly referenced. 
    \end{itemize}

    \item {\bf Experimental result reproducibility}
    \item[] Question: Does the paper fully disclose all the information needed to reproduce the main experimental results of the paper to the extent that it affects the main claims and/or conclusions of the paper (regardless of whether the code and data are provided or not)?
    \item[] Answer: \answerYes{} 
    \item[] Justification: The paper provides substantial details needed to reproduce the main results, including model architectures, loss definitions, training objectives, dataset mixture, sampling strategy, optimization hyperparameters, benchmark protocols, and evaluation settings; see Secs.~\ref{sec:pretraining}, \ref{sec:transfer}, and~\ref{sec:exp_setup}, and Appendices~\ref{app:impl}, \ref{app:downstream}, \ref{app:eval}, and~\ref{app:real_world}.
    \item[] Guidelines:
    \begin{itemize}
        \item The answer \answerNA{} means that the paper does not include experiments.
        \item If the paper includes experiments, a \answerNo{} answer to this question will not be perceived well by the reviewers: Making the paper reproducible is important, regardless of whether the code and data are provided or not.
        \item If the contribution is a dataset and\slash or model, the authors should describe the steps taken to make their results reproducible or verifiable. 
        \item Depending on the contribution, reproducibility can be accomplished in various ways. For example, if the contribution is a novel architecture, describing the architecture fully might suffice, or if the contribution is a specific model and empirical evaluation, it may be necessary to either make it possible for others to replicate the model with the same dataset, or provide access to the model. In general. releasing code and data is often one good way to accomplish this, but reproducibility can also be provided via detailed instructions for how to replicate the results, access to a hosted model (e.g., in the case of a large language model), releasing of a model checkpoint, or other means that are appropriate to the research performed.
        \item While NeurIPS does not require releasing code, the conference does require all submissions to provide some reasonable avenue for reproducibility, which may depend on the nature of the contribution. For example
        \begin{enumerate}
            \item If the contribution is primarily a new algorithm, the paper should make it clear how to reproduce that algorithm.
            \item If the contribution is primarily a new model architecture, the paper should describe the architecture clearly and fully.
            \item If the contribution is a new model (e.g., a large language model), then there should either be a way to access this model for reproducing the results or a way to reproduce the model (e.g., with an open-source dataset or instructions for how to construct the dataset).
            \item We recognize that reproducibility may be tricky in some cases, in which case authors are welcome to describe the particular way they provide for reproducibility. In the case of closed-source models, it may be that access to the model is limited in some way (e.g., to registered users), but it should be possible for other researchers to have some path to reproducing or verifying the results.
        \end{enumerate}
    \end{itemize}

\item {\bf Open access to data and code}
    \item[] Question: Does the paper provide open access to the data and code, with sufficient instructions to faithfully reproduce the main experimental results, as described in supplemental material?
    \item[] Answer: \answerYes{} 
    \item[] Justification: Our code is available at: \textit{https://github.com/Mark-zjtang/ALAM/tree/main}
    \item[] Guidelines:
    \begin{itemize}
        \item The answer \answerNA{} means that paper does not include experiments requiring code.
        \item Please see the NeurIPS code and data submission guidelines (\url{https://neurips.cc/public/guides/CodeSubmissionPolicy}) for more details.
        \item While we encourage the release of code and data, we understand that this might not be possible, so \answerNo{} is an acceptable answer. Papers cannot be rejected simply for not including code, unless this is central to the contribution (e.g., for a new open-source benchmark).
        \item The instructions should contain the exact command and environment needed to run to reproduce the results. See the NeurIPS code and data submission guidelines (\url{https://neurips.cc/public/guides/CodeSubmissionPolicy}) for more details.
        \item The authors should provide instructions on data access and preparation, including how to access the raw data, preprocessed data, intermediate data, and generated data, etc.
        \item The authors should provide scripts to reproduce all experimental results for the new proposed method and baselines. If only a subset of experiments are reproducible, they should state which ones are omitted from the script and why.
        \item At submission time, to preserve anonymity, the authors should release anonymized versions (if applicable).
        \item Providing as much information as possible in supplemental material (appended to the paper) is recommended, but including URLs to data and code is permitted.
    \end{itemize}

\item {\bf Experimental setting/details}
    \item[] Question: Does the paper specify all the training and test details (e.g., data splits, hyperparameters, how they were chosen, type of optimizer) necessary to understand the results?
    \item[] Answer: \answerYes{} 
    \item[] Justification: The paper specifies the main training and evaluation details, including architectures, data splits, hyperparameters, optimizers, action horizons, pretraining data mixture, benchmark protocols, and real-world evaluation setup; see Sec.~\ref{sec:exp_setup} and Appendices~\ref{app:impl}, \ref{app:downstream}, \ref{app:eval}, and~\ref{app:real_world}.
    \item[] Guidelines:
    \begin{itemize}
        \item The answer \answerNA{} means that the paper does not include experiments.
        \item The experimental setting should be presented in the core of the paper to a level of detail that is necessary to appreciate the results and make sense of them.
        \item The full details can be provided either with the code, in appendix, or as supplemental material.
    \end{itemize}

\item {\bf Experiment statistical significance}
    \item[] Question: Does the paper report error bars suitably and correctly defined or other appropriate information about the statistical significance of the experiments?
    \item[] Answer: \answerNo{} 
    \item[] Justification: The paper reports averaged success rates and evaluation counts, and notes statistical noise for a small task tier in Appendix~\ref{app:latent_intervention}, but it does not report confidence intervals, standard deviations across random seeds, or formal statistical significance tests for the main results.
    \item[] Guidelines:
    \begin{itemize}
        \item The answer \answerNA{} means that the paper does not include experiments.
        \item The authors should answer \answerYes{} if the results are accompanied by error bars, confidence intervals, or statistical significance tests, at least for the experiments that support the main claims of the paper.
        \item The factors of variability that the error bars are capturing should be clearly stated (for example, train/test split, initialization, random drawing of some parameter, or overall run with given experimental conditions).
        \item The method for calculating the error bars should be explained (closed form formula, call to a library function, bootstrap, etc.)
        \item The assumptions made should be given (e.g., Normally distributed errors).
        \item It should be clear whether the error bar is the standard deviation or the standard error of the mean.
        \item It is OK to report 1-sigma error bars, but one should state it. The authors should preferably report a 2-sigma error bar than state that they have a 96\% CI, if the hypothesis of Normality of errors is not verified.
        \item For asymmetric distributions, the authors should be careful not to show in tables or figures symmetric error bars that would yield results that are out of range (e.g., negative error rates).
        \item If error bars are reported in tables or plots, the authors should explain in the text how they were calculated and reference the corresponding figures or tables in the text.
    \end{itemize}

\item {\bf Experiments compute resources}
    \item[] Question: For each experiment, does the paper provide sufficient information on the computer resources (type of compute workers, memory, time of execution) needed to reproduce the experiments?
    \item[] Answer: \answerYes{} 
    \item[] Justification: The paper reports the main compute resources used for pretraining and downstream finetuning, including GPU type, GPU count, memory, and approximate runtime; see Sec.~\ref{sec:exp_setup} and Appendices~\ref{app:impl}.
    \item[] Guidelines:
    \begin{itemize}
        \item The answer \answerNA{} means that the paper does not include experiments.
        \item The paper should indicate the type of compute workers CPU or GPU, internal cluster, or cloud provider, including relevant memory and storage.
        \item The paper should provide the amount of compute required for each of the individual experimental runs as well as estimate the total compute. 
        \item The paper should disclose whether the full research project required more compute than the experiments reported in the paper (e.g., preliminary or failed experiments that didn't make it into the paper). 
    \end{itemize}
    
\item {\bf Code of ethics}
    \item[] Question: Does the research conducted in the paper conform, in every respect, with the NeurIPS Code of Ethics \url{https://neurips.cc/public/EthicsGuidelines}?
    \item[] Answer: \answerYes{} 
    \item[] Justification: The authors have reviewed the NeurIPS Code of Ethics and, to the best of our knowledge, the research conforms to it. The work uses standard robot-learning datasets and benchmarks and does not involve deceptive practices or human-subject experimentation.
    \item[] Guidelines:
    \begin{itemize}
        \item The answer \answerNA{} means that the authors have not reviewed the NeurIPS Code of Ethics.
        \item If the authors answer \answerNo, they should explain the special circumstances that require a deviation from the Code of Ethics.
        \item The authors should make sure to preserve anonymity (e.g., if there is a special consideration due to laws or regulations in their jurisdiction).
    \end{itemize}

\item {\bf Broader impacts}
    \item[] Question: Does the paper discuss both potential positive societal impacts and negative societal impacts of the work performed?
    \item[] Answer: \answerYes{} 
    \item[] Justification: The paper discusses broader impacts in Appendix~\ref{app:broader_impact}, including the potential to reduce reliance on action-labeled robot data and the need for appropriate evaluation and safety measures in real-world deployment.
    \item[] Guidelines:
    \begin{itemize}
        \item The answer \answerNA{} means that there is no societal impact of the work performed.
        \item If the authors answer \answerNA{} or \answerNo, they should explain why their work has no societal impact or why the paper does not address societal impact.
        \item Examples of negative societal impacts include potential malicious or unintended uses (e.g., disinformation, generating fake profiles, surveillance), fairness considerations (e.g., deployment of technologies that could make decisions that unfairly impact specific groups), privacy considerations, and security considerations.
        \item The conference expects that many papers will be foundational research and not tied to particular applications, let alone deployments. However, if there is a direct path to any negative applications, the authors should point it out. For example, it is legitimate to point out that an improvement in the quality of generative models could be used to generate Deepfakes for disinformation. On the other hand, it is not needed to point out that a generic algorithm for optimizing neural networks could enable people to train models that generate Deepfakes faster.
        \item The authors should consider possible harms that could arise when the technology is being used as intended and functioning correctly, harms that could arise when the technology is being used as intended but gives incorrect results, and harms following from (intentional or unintentional) misuse of the technology.
        \item If there are negative societal impacts, the authors could also discuss possible mitigation strategies (e.g., gated release of models, providing defenses in addition to attacks, mechanisms for monitoring misuse, mechanisms to monitor how a system learns from feedback over time, improving the efficiency and accessibility of ML).
    \end{itemize}
    
\item {\bf Safeguards}
    \item[] Question: Does the paper describe safeguards that have been put in place for responsible release of data or models that have a high risk for misuse (e.g., pre-trained language models, image generators, or scraped datasets)?
    \item[] Answer: \answerNA{} 
    \item[] Justification: The paper does not release a high-risk generative model or scraped dataset requiring special safeguards. The method is evaluated as a robotics policy-learning component under standard research benchmarks and controlled robot settings.
    \item[] Guidelines:
    \begin{itemize}
        \item The answer \answerNA{} means that the paper poses no such risks.
        \item Released models that have a high risk for misuse or dual-use should be released with necessary safeguards to allow for controlled use of the model, for example by requiring that users adhere to usage guidelines or restrictions to access the model or implementing safety filters. 
        \item Datasets that have been scraped from the Internet could pose safety risks. The authors should describe how they avoided releasing unsafe images.
        \item We recognize that providing effective safeguards is challenging, and many papers do not require this, but we encourage authors to take this into account and make a best faith effort.
    \end{itemize}

\item {\bf Licenses for existing assets}
    \item[] Question: Are the creators or original owners of assets (e.g., code, data, models), used in the paper, properly credited and are the license and terms of use explicitly mentioned and properly respected?
    \item[] Answer: \answerNo{} 
    \item[] Justification: The paper credits the original sources of the datasets, backbones, and benchmarks used throughout the experiments via citations, but it does not explicitly enumerate the licenses and terms of use of each external asset in the manuscript or appendix.
    \item[] Guidelines:
    \begin{itemize}
        \item The answer \answerNA{} means that the paper does not use existing assets.
        \item The authors should cite the original paper that produced the code package or dataset.
        \item The authors should state which version of the asset is used and, if possible, include a URL.
        \item The name of the license (e.g., CC-BY 4.0) should be included for each asset.
        \item For scraped data from a particular source (e.g., website), the copyright and terms of service of that source should be provided.
        \item If assets are released, the license, copyright information, and terms of use in the package should be provided. For popular datasets, \url{paperswithcode.com/datasets} has curated licenses for some datasets. Their licensing guide can help determine the license of a dataset.
        \item For existing datasets that are re-packaged, both the original license and the license of the derived asset (if it has changed) should be provided.
        \item If this information is not available online, the authors are encouraged to reach out to the asset's creators.
    \end{itemize}

\item {\bf New assets}
    \item[] Question: Are new assets introduced in the paper well documented and is the documentation provided alongside the assets?
    \item[] Answer: \answerNA{} 
    \item[] Justification: The paper does not currently release a new public dataset, codebase, or model asset as part of the submission.
    \item[] Guidelines:
    \begin{itemize}
        \item The answer \answerNA{} means that the paper does not release new assets.
        \item Researchers should communicate the details of the dataset\slash code\slash model as part of their submissions via structured templates. This includes details about training, license, limitations, etc. 
        \item The paper should discuss whether and how consent was obtained from people whose asset is used.
        \item At submission time, remember to anonymize your assets (if applicable). You can either create an anonymized URL or include an anonymized zip file.
    \end{itemize}

\item {\bf Crowdsourcing and research with human subjects}
    \item[] Question: For crowdsourcing experiments and research with human subjects, does the paper include the full text of instructions given to participants and screenshots, if applicable, as well as details about compensation (if any)? 
    \item[] Answer: \answerNA{} 
    \item[] Justification: The paper does not involve crowdsourcing experiments or research with human subjects. The demonstrations used for real-world robot training are teleoperated robot demonstrations rather than a human-subject study.
    \item[] Guidelines:
    \begin{itemize}
        \item The answer \answerNA{} means that the paper does not involve crowdsourcing nor research with human subjects.
        \item Including this information in the supplemental material is fine, but if the main contribution of the paper involves human subjects, then as much detail as possible should be included in the main paper. 
        \item According to the NeurIPS Code of Ethics, workers involved in data collection, curation, or other labor should be paid at least the minimum wage in the country of the data collector. 
    \end{itemize}

\item {\bf Institutional review board (IRB) approvals or equivalent for research with human subjects}
    \item[] Question: Does the paper describe potential risks incurred by study participants, whether such risks were disclosed to the subjects, and whether Institutional Review Board (IRB) approvals (or an equivalent approval/review based on the requirements of your country or institution) were obtained?
    \item[] Answer: \answerNA{} 
    \item[] Justification: The paper does not involve crowdsourcing or human-subject research requiring IRB review. No personal or sensitive human data is collected or analyzed.
    \item[] Guidelines:
    \begin{itemize}
        \item The answer \answerNA{} means that the paper does not involve crowdsourcing nor research with human subjects.
        \item Depending on the country in which research is conducted, IRB approval (or equivalent) may be required for any human subjects research. If you obtained IRB approval, you should clearly state this in the paper. 
        \item We recognize that the procedures for this may vary significantly between institutions and locations, and we expect authors to adhere to the NeurIPS Code of Ethics and the guidelines for their institution. 
        \item For initial submissions, do not include any information that would break anonymity (if applicable), such as the institution conducting the review.
    \end{itemize}

\item {\bf Declaration of LLM usage}
    \item[] Question: Does the paper describe the usage of LLMs if it is an important, original, or non-standard component of the core methods in this research? Note that if the LLM is used only for writing, editing, or formatting purposes and does \emph{not} impact the core methodology, scientific rigor, or originality of the research, declaration is not required.
    \item[] Answer: \answerNA{} 
    \item[] Justification: The core methodological contribution of this work is ALAM and its transfer mechanism, not a new LLM-based method. Existing PaliGemma/Gemma backbones are used only as standard downstream VLA components, as described in Sec.~\ref{sec:exp_setup} and Appendix~\ref{app:downstream}.
    \item[] Guidelines:
    \begin{itemize}
        \item The answer \answerNA{} means that the core method development in this research does not involve LLMs as any important, original, or non-standard components.
        \item Please refer to our LLM policy in the NeurIPS handbook for what should or should not be described.
    \end{itemize}

\end{enumerate}

\end{document}